\documentclass{article}

\usepackage{arxiv}

\usepackage[utf8]{inputenc} 
\usepackage[T1]{fontenc}    
\usepackage{hyperref}       
\usepackage{url}            
\usepackage{booktabs}       
\usepackage{amsfonts}       
\usepackage{nicefrac}       
\usepackage{microtype}      
\usepackage{lipsum}
\usepackage{graphicx}
\usepackage{algorithm}
\usepackage{subfigure}
\usepackage{amsmath}

\title{Markerless Stride Length estimation in Athletic  using Pose Estimation with monocular vision}

\author{
 Patryk Skorupski \\
  Dipartimento di Ingegneria dell'Innovazione\\
  Università del Salento\\
  Campus Universitario, Via Monteroni sn,  73100 Lecce, (ITALY) \\
  \texttt{patryk.skorupski@studenti.unisalento.it} \\
   \And
 Cosimo Distante \\
  Istituto di Scienze Applicate e sIstemi Intelligenti - ISASI\\
 Consiglio Nazionale delle Ricerche - (CNR)\\
  c/o DHITECH -   Campus Universitario, Via Monteroni sn. 73100 Lecce, (ITALY)   \\
  \texttt{cosimo.distante@cnr.it} \\
  \And
 Pier Luigi Mazzeo \\
 Istituto di Scienze Applicate e sIstemi Intelligenti - ISASI\\
 Consiglio Nazionale delle Ricerche - (CNR)\\
  c/o DHITECH -   Campus Universitario, Via Monteroni sn, 73100 Lecce, (ITALY)   \\
  \texttt{pierluigi.mazzeo@cnr.it} \\
}

\begin{document}
\maketitle
\begin{abstract}
Performance measures such as stride length in athletics and the pace of runners can be estimated using different tricks such as measuring the number of steps divided by the running length or helping with markers printed on the track.    
Monitoring individual performance is essential for supporting staff coaches in establishing a proper training schedule for each athlete.  The aim of this paper is to investigate a computer vision-based approach for estimating stride length and speed transition from video sequences and assessing video analysis processing among athletes. Using some well-known image processing methodologies such as probabilistic hough transform combined with a human pose detection algorithm, we estimate the leg joint position of runners. In this way, applying a homography transformation, we can estimate the runner stride length. Experiments on various race videos with three different runners demonstrated that the proposed system represents a useful tool for coaching and training. This suggests its potential value in measuring and monitoring the gait parameters of athletes.
\end{abstract}



\section{Introduction} \label{sec:Intro}
Track and field is one of the most practiced sports with a profound historical background and highly competitive nature. This discipline demands exceptionally high standards for athletes' technical skill and physical condition \cite{Lu_2025} \cite{Zhu2022}. In the past, coaches' experience and intuition have been the foundational components of track and field training techniques. While this approach has its advantages, it frequently lacks objective measures for assessing the precision and depth of technical actions.  As a result, this dependence on personal opinion may bring an inaccurate and inadequate assessment of athletes' performances \cite{mingtaou2023study} \cite{zheng2022improved}. In track and field, traditional evaluation methods are usually limited by their incapacity to produce thorough and measurable data.5, 6 This oversight highlights the need for more reliable evaluation techniques that can provide accurate and useful information about athletes' overall performance and technical execution. Deep learning models and machine vision technology have opened up new avenues for improving evaluation efficiency and accuracy. A model based on pose estimation methodologies allows the tracking of key body joints and movements, extracting valuable data for analyzing athletic performance \cite{wang2019human} \cite{cronin2024feasibility}. 
Despite these developments, there is still a lot of space for improvement, especially in the areas of accuracy, real-time processing, and specific training adaptability. Some of the gaps left by traditional evaluation techniques have been filled in large part by machine vision technology. These technologies are able to analyze complex movements more accurately by capturing detailed visual data of athletes in action. This capability has been enhanced by the integration of deep learning models, which make it possible to identify and interpret patterns in the visual data automatically. These models process huge amounts of data,  which gradually increase the precision of their predictions and analyses by learning from previous data examples \cite{archana2024deep} \cite{Detect2Interact}. Using a pose estimation model such as EfficientPose \cite{zhang2021efficientpose} and  OpenPose \cite{OpenPose} in track and field training sessions can lead to more accurate and faster assessments of runners' performance. The integration of these computer vision models allows useful metrics from runners to be extracted, including split times, split distances, step length, and ground context time. These measures have been utilized in different coaching practices in order to enhance athletes' performance in both training and competitive settings. 
In this paper, we propose an approach to measuring athletes' movements using runners' videos. Its main contributions consist of: i) combining classical image processing (Canny edge detection, Probabilistic Hough Transform) with modern deep learning for markerless stride length estimation; ii) using homography to transform 2D video frames into a measurable top-down view, enabling metric distance calculations; and iii) addressing a real-world need in sports analytics by providing a tool for coaches to monitor athletes' performance without intrusive markers. It demonstrates feasibility with experiments on three runners, showing consistent stride length measurements across videos, making the system suitable for live training sessions.
The remainder of the paper includes four sections. Section \ref{sec:Mat&Meth} in which are presented the methodology and the material used. Section  \ref{sec:experiments} describes how the experiments has been conducted. Finally section \ref{sec:conclusion} summarizes and comments obtained results and addresses future work.

\section{Materials and Method} \label{sec:Mat&Meth}
\begin{figure}
\includegraphics[width=\textwidth]{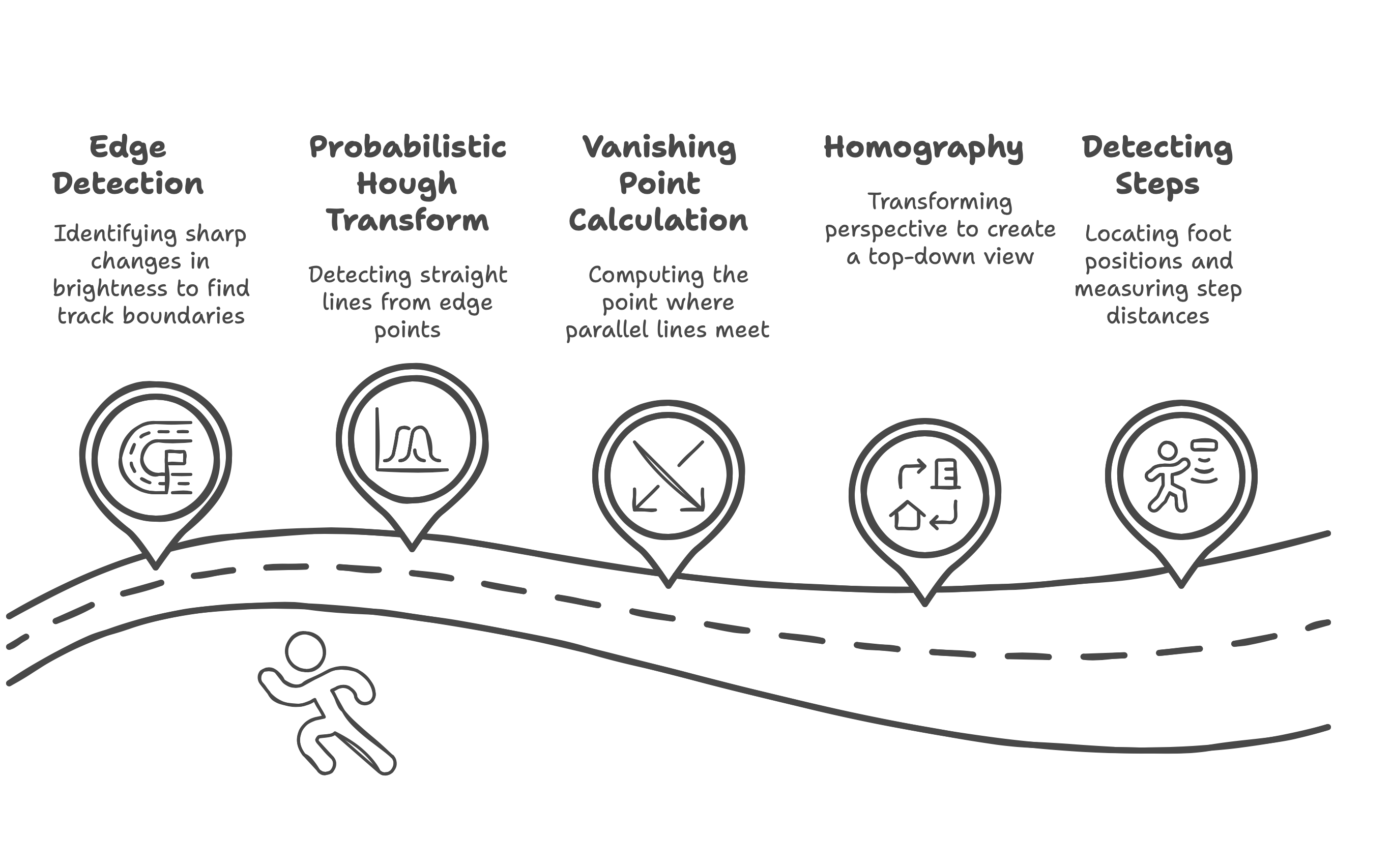}
\caption{Pipeline of the proposed methodology} \label{fig:Method_Overview}
\end{figure}

Our method can be divided into different steps as schematized in figure \ref{fig:Method_Overview}. 
\textbf{Edge Detection:} Analyzing the image to detect sharp changes in brightness (edges), which helps identify the boundaries of the track or lane lines. In this way, we can isolate the most prominent lines in the scene for further analysis.
\textbf{Probabilistic Hough Transform:} A  voting algorithm that detects straight lines from edge points, including their start/end positions and intersections. It allows us to extract the dominant lines (e.g., lane markings) and detects where they cross or converge.
\textbf{Vanishing Point Calculation:} For computing the point where parallel lines in the scene (like track boundaries) appear to meet in the image due to perspective. Using this information helps us to estimate the depth and orientation of the track relative to the camera. 
\textbf{Homography}: Applying a geometric transformation (homography matrix) to warp the perspective and create a top-down view of the track ('bird's eye'). It converts the skewed camera view into a measurable 2D plane for accurate distance calculations.
Finally, \textbf{Detecting Steps:} Using EfficientPose \cite{zhang2021efficientpose} (a pose/keypoint detection method) to locate foot positions and measure the distance between consecutive steps. It quantifies gait parameters (e.g., stride length) for biomechanical analysis.

\subsection{Edge Detection} 
\begin{figure}
\includegraphics[width=\textwidth]{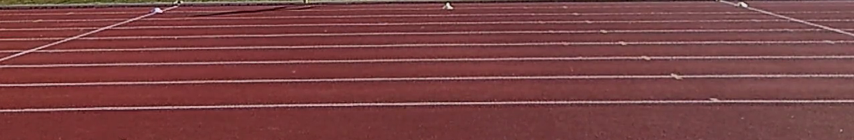}
\caption{Training area which contains the lines.} \label{fig:ROI}
\end{figure}

\begin{figure}
\includegraphics[width=\textwidth]{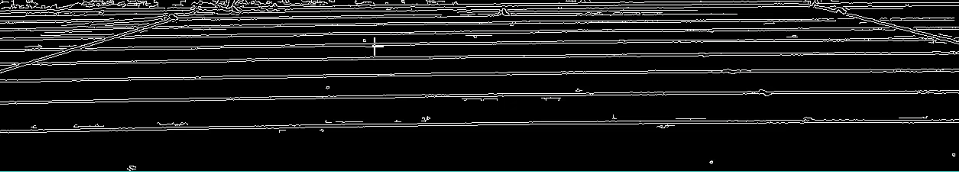}
\caption{Edge extracted from the track.} \label{fig:Edge}
\end{figure}

The region of interest (ROI) for geometric analysis and line identification is strictly limited to the track itself. Extraneous regions of the scene have been deliberately excluded from processing to mitigate the introduction of interfering artifacts and noise, which could degrade the precision of line detection algorithms. To obtain sharper edges and reduce noise, the scene was first converted to grayscale, followed by the application of a Gaussian filter. For edge detection, the Canny method has been employed. Figure \ref{fig:ROI} contains the ROI regions with track lines must be detected, instead Figure \ref{fig:Edge} showed the results ofter the Canny method application.

\subsection{Probabilistic Hough Transform} For track line detection, we employed the Probabilistic Hough Transform (PHT), which, unlike the standard Hough Transform, returns finite line segments rather than infinite lines. This approach proves particularly advantageous in our context, as PHT enables detection of shorter line segments with higher precision - a critical aspect for our objectives.
Why is precision crucial? For the aim of this paper is essential the geometric rectification of the scene, which requires accurate determination of intersection coordinates between the track's vertical and horizontal lines. Precise line detection is fundamental to ensure a Correct homography transformation and an Accurate derived measurements (e.g., stride length). The PHT's segment-based detection provides the necessary spatial accuracy for these computational geometry operations. 
The principal strenghts of the PHT are summarized in the follow three point:
\begin{itemize}
    \item Finite Segments: More practical for identifying points of interest compared to infinite lines, particularly given the video geometry containing slightly curved trajectories.
    \item Precise Segment Control: PHT provides tunable parameters like minLineLength and maxLineGap, enabling rigorous filtering of line segments to optimize detection accuracy.
    \item Computational Efficiency: PHT demonstrates superior runtime performance relative to the classical Hough Transform, making it better suited for real-time applications.
\end{itemize}


\subsection{Detecting Lines}
We need to correctly identify each track lines, for doing that we need to execute three consecutive steps:
\begin{itemize}
    \item Horizontal lines detection and identification
    \item Vertical lines  detection and identification
    \item Vanishing point estimation
\end{itemize}
\subsubsection{Horizontal lines detection and identification}
The first aim of this step is to precisely detect the horizontal track lines to accurately identify their intersection points with the vertical lines. For doing that  we use a PHT thresholds (minLineLength, maxLineGap) for horizontal line filtering 
Angle constraints (e.g., ±10° tolerance for horizontal line classification).

\begin{figure}[H]
    \includegraphics[width=\textwidth]{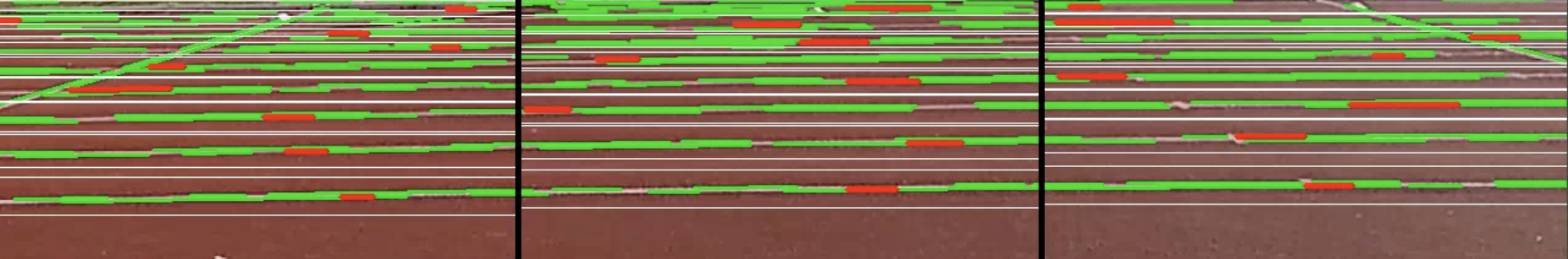}
    \caption{Segments grouped by area, the reference segment in red, the thresholds for each of the three areas in white} \label{fig:SegmentArea}
    \end{figure}
\begin{figure}[H]
    \includegraphics[width=\textwidth]{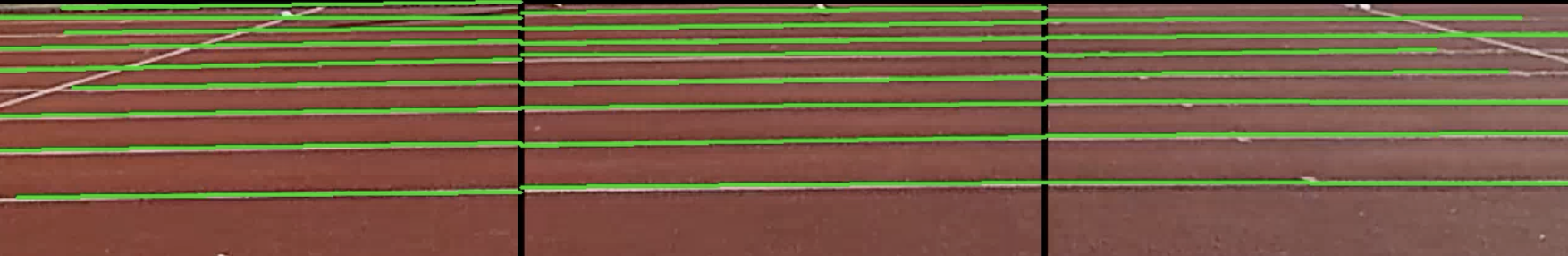}
    \caption{Merged segments grouped by area} \label{fig:FullSegmentArea}
\end{figure}

To handle the natural curvature of horizontal track lines, we implement a four-stage adaptive grouping algorithm that:
\begin{enumerate}
    \item Segment lines were further divided into three smaller sections, in order to facilitate their processing. This subdivision
    allowed for more accurate grouping and merging of the lines, especially in curved
    areas where a global model would have failed to correctly represent the geometry.
    \item The track was
    divided into three distinct areas (left/mid/right), each relating to a smaller portion of the overall
    scene. 
    \item A variable threshold value was adopted to the distance from
    the camera. Since the runway lines appear closer together as they move away
    from each other in the image perspective, it was chosen to reduce the threshold
    value in the more distant regions, in order to capture these denser features more
    accurately figure \ref{fig:SegmentArea}.

    \item For each area and for each threshold value, a reference line representative of the group was selected. Subsequently, all segments falling within the defined threshold were merged into a single final line. The fusion criterion adopted was based on the horizontal extension of the segments: the resulting line was defined such that the value \( x_1 \) corresponds to the minimum \( x \) value among all the segments in the subarea, while \( x_2 \) corresponds to the maximum \( x \) value. The values \( y_1 \) and \( y_2 \) were taken from the endpoints associated with these \( x \) values.
    
    This approach ensured that the final line obtained covered the entire portion of the visible subsection as in the figure \ref{fig:FullSegmentArea}.

    \item Finally, to join the lines belonging to adjacent areas, Euclidean distance rules were applied between the most extreme points of the segments obtained in point four.
    The pseudocode that summarizes the logical steps described above is contained in  the Algorithm \ref{alg:AdaptiveLine}.
\end{enumerate}

\begin{figure}
\includegraphics[width=\textwidth]{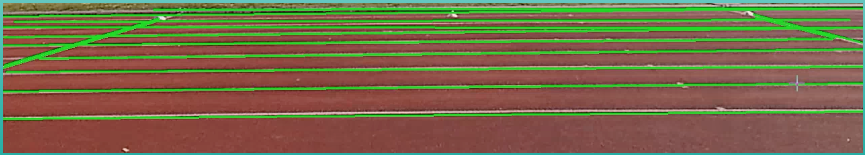}
\caption{Vertical and Horizontal lines detected by the proposed method.} \label{fig:lines_final}
\end{figure}

Figure \ref{fig:lines_final} contains the final results  of the described track line detection algorithm. While distant lines appear unstable, this does not affect our purpose, since only four reference points—derived from the intersections of the fourth and fifth horizontal lines with the vertical lines—are required for homography estimation.

\subsubsection{Vertical lines  detection and identification}

Vertical lines were identified using a threshold of 2 degrees. All lines with an angle exceeding this value relative to the base angle were classified as vertical.
Selection of Dominant Vertical Lines: Vertical lines were grouped based on their angle, with a threshold of 15 degrees. Among similar lines, the longest one was selected.
Notice that if "inclinate" was intended to mean "near-vertical" rather than strictly vertical, the translation can be adjusted to: "Identification of Near-Vertical Lines" and "Selection of Dominant Near-Vertical Lines." Clarify if a different interpretation is needed.

\subsubsection{Vanishing point estimation}
\begin{figure}
\includegraphics[width=\textwidth]{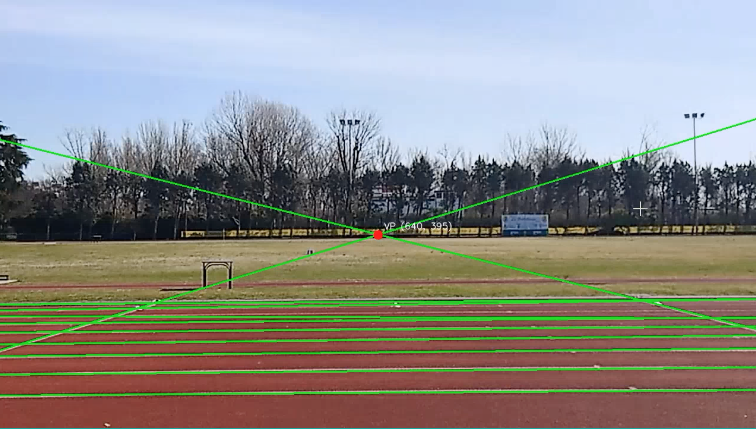}
\caption{Vanishing point detection using identified vertical lines.} \label{fig:vanishing_point}
\end{figure}

The lines were extrapolated over the entire image by computing their slope and intercept.
The vanishing point was derived as the intersection point of the two vertical lines.
Figure \ref{fig:vanishing_point} show how the vanishing point has been estimated.

\subsection{Homography}
The computation of a planar homography requires the establishment of at least four point correspondences between two distinct planes: the projective image plane $\mathbb{P}^2$(pixel coordinates) and the Euclidean world plane $\mathbb{R}^2$ (metric coordinates). This minimal configuration is mathematically necessary due to the properties of the homography transformation in projective geometry. The computation of a planar homography requires establishing a minimum of four corresponding point pairs between two distinct coordinate systems: the image plane (2D pixel coordinates) and the world plane (metric coordinates). This fundamental requirement stems from the mathematical properties of projective transformations:
The homography matrix $H\in \mathbb{R}^{3\times 3}$  possesses 8 degrees of freedom (up to scale )
Each point correspondence provides two independent linear equations through the transformation:

\begin{equation}
    \begin{bmatrix} 
        x' \\ 
        y' \\ 
        w' 
    \end{bmatrix} 
    = \mathbf{H} 
    \begin{bmatrix} 
        x \\ 
        y \\ 
        1 
    \end{bmatrix}
    \label{eq:homography}
\end{equation}
where $(x,y)$ are image coordinates and $(x'/w', y'/w')$ are world coordinates.

\subsection{Detecting Steps}
\begin{figure}
\centering
\includegraphics[width=0.75\textwidth]{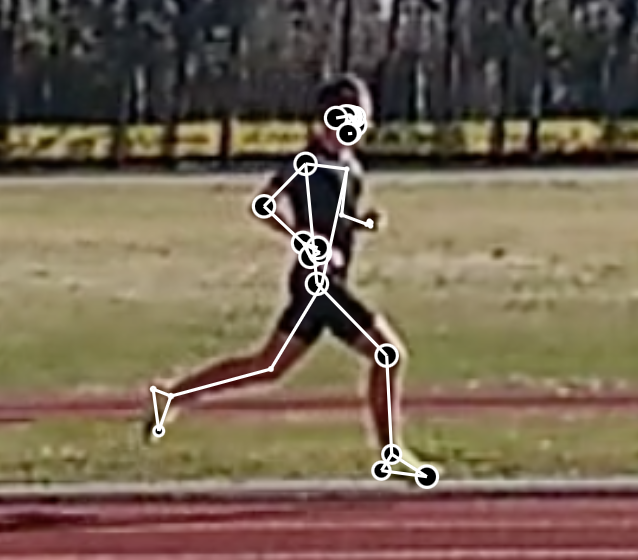}
\caption{The white circle identified the  detected landmarks for Athlete pose estimation. The  33 landmarks represent various distinctive parts of the human body.} \label{fig:DetectedKeyPoints}
\end{figure}

\begin{figure}
\includegraphics[width=\textwidth]{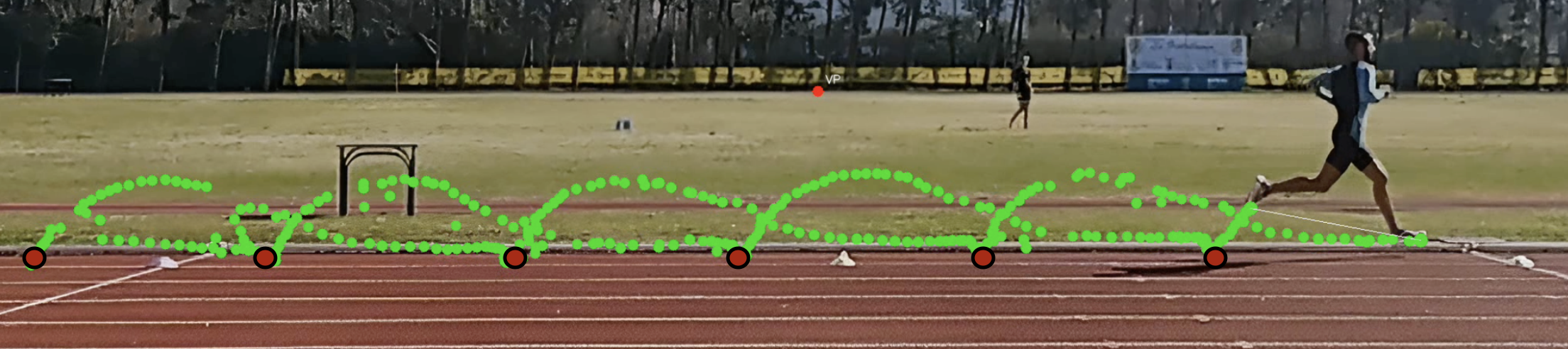}
\caption{In yellow all the key points detected while in red those considered for the distance calculation.} \label{fig:KeyPoints}
\end{figure}

For the Pose Estimation method we have evaluated  EfficientPose \cite{zhang2021efficientpose}, OpenPose and MoveNet  \cite{MoveNet}. EfficientPose achieves better performance for one-person 2D analysis resulting much faster than OpenPose. Instead MoveNet resulted optimized for mobile devices but lacks in 3D support. We choose EfficientPose for Human Posture Estimation because it is optimized for real-time performance while maintaining accuracy. It uses EfficientNet backbones with multi-task learning to predict 2D/3D keypoints, body orientation, and bounding boxes. The model scales across variants (I-IV) for speed-accuracy trade-offs, achieving 30+ FPS on consumer GPUs. Unlike traditional top-down approaches, it efficiently handles occlusions and multi-person scenarios with a single forward pass. It's ideal for applications like AR, sports analytics, and healthcare monitoring, balancing speed and precision better than RCNN-based or PnP methods. EfficientPose excels in 3D-aware applications needing real-time performance.  Figure \ref{fig:DetectedKeyPoints} describes the output of the EfficientPose model that predicts a full set of 3D pose landmarks. With this pose detection model is possible to correct identify body locations and key points.  Furthermore the pose landmarker model refining this into a detailed 3D map of 33 anatomical landmarks as the figure \ref{fig:DetectedKeyPoints} showed. 
In our method, we used the positions of the toes as keypoints to determine the foot landing points. To determine the exact point of contact between the track and the foot, the concept of keypoint velocity was used, if the keypoint remained stationary in the same pixel for two consecutive frames, it was considered a landing point figure \ref{fig:KeyPoints}.

\section{Experiments} \label{sec:experiments}
\begin{figure}
\begin{subfigure}{}
    \includegraphics[width=\textwidth]{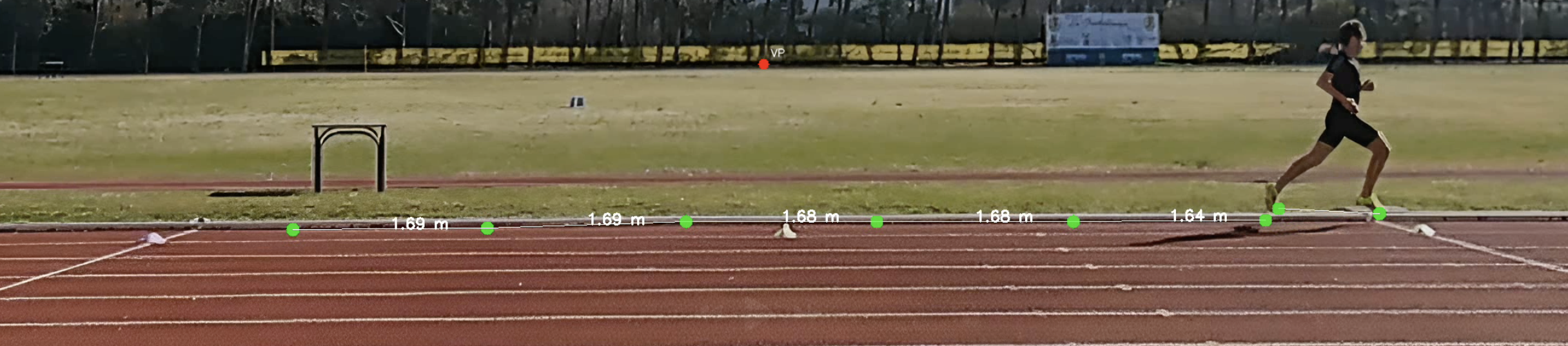}
    \caption{First Runner.}
    \label{fig:FinalResult1}
\end{subfigure}

\vspace{1cm} 

\begin{subfigure}{}
    \includegraphics[width=\textwidth]{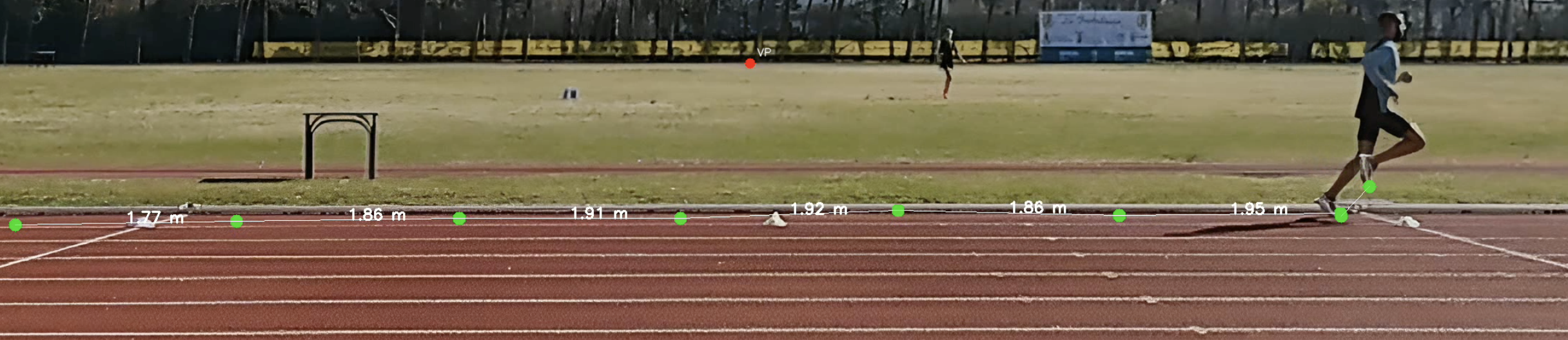}
    \caption{Second Runner.}
    \label{fig:FinalResult2}
\end{subfigure}

\caption{Final result on stride length calculation for two different runners.}
\label{fig:FinalResult}
\end{figure}

A high-definition camera has been installed near the training pitch.  The position of the camera allow us to acquire a frontal view of the track as shown in figure \ref{fig:vanishing_point}. The camera can provide high-quality image data, 1280$\times$720 pixels, 30 frames per second,  that allows the extraction of information about the athlete's posture, movement speed and angle.%
Videos were processed in their entirety, performing line detection on each frame and computing the homography. For each frame, the homography matrix was calculated three times, considering the intersections between the vertical lines and the first two lines, the second and third lines, and finally the fourth and fifth lines.
To obtain the final homography matrix, the median of the matrices computed for each individual frame was taken to mitigate outliers caused by erroneous line detection and noise. After identifying the key points using the stationarity-based technique (described above) each point was projected along the line formed by the point itself and the vanishing point in the direction of the less noisy scene. This allows to obtain a more reliable position for the distance calculation. The previously calculated homography was applied to transform the points from the image plane to the real plane. Finally, the distances between the support points were estimated using the Euclidean distance in the transformed plane, thus obtaining an approximate but consistent measure of the step length figure \ref{fig:FinalResult}. The table \ref{tab:risultati} shows the average stride length for three runners, based on two separate videos for each athlete. 
These measures show some variability, which could be due to several factors. Since there is no ground truth (i.e., no objectively measured "true" stride length), and  considering that the foot position is ever correctly detected for each steps,  we can only compare the consistency between the two videos and discuss possible reasons for the differences. Athlete 1: Stride length increased from 1.66 m (Video 1) to 1.76 m (Video 2) (+0.10 m difference). This is a noticeable increase (~6 \% change), which could suggest differences in running speed and fatigue. Athlete 2: Stride length decreased slightly from 1.87 m (Video 1) to 1.80 m (Video 2) (-0.07 m difference). A small decrease (~4\%), possibly due to pacing adjustments. Athlete 3: Stride length decreased slightly from 1.90 m (Video 1) to 1.86 m (Video 2) (-0.04 m difference). A very small change (~2\%), suggesting relatively consistent stride length between videos.  Such variability is due to: i) running Conditions, athletes may have been running at different speeds (stride length typically increases with speed); ii) Fatigue or pacing conditions could affect stride dynamics. 
Even at the same speed, the stride length can vary slightly due to terrain, footwear, or biomechanical adjustments.

\begin{table}[h]
\centering
\begin{tabular}{|c|c|c|}
\hline
         & Average stride length Video 1 & Average stride length Video 2 \\
\hline
Athlete 1 & 1.66 m & 1.76 m \\
\hline
Athlete 2 & 1.87 m & 1.80 m \\
\hline
Athlete 3 & 1.90 m & 1.86 m \\
\hline
\end{tabular}
\caption{Average stride length measured for each athlete in two separate videos.}
\label{tab:risultati}
\end{table}

\section{Conclusion} \label{sec:conclusion}
This paper demonstrates the feasibility and effectiveness of a computer vision-based approach to predicting a runner's stride from simple videos. Even though ground truth data is not available, the measured stride lengths appear consistent and exhibit minimal variation for each athlete between the two video recordings. The approach demonstrated that excellent results can be achieved using techniques based on classical image processing systems, combined with more advanced pose detection models such as EfficientPose. Accurate line detection resulted in a homography matrix that could convert 2D coordinates into real metric values, allowing for a reliable estimate of step length.
Additionally, foot contact point detection, based on keypoint velocity analysis, ensures consistency and robustness even in dynamic conditions. The results obtained confirm the reliability of the method and its applicability in training and competitions, to constantly monitor the athletes' stride. Future work will include: 1) compare results with ground truth data (e.g., motion capture systems or force plates) to quantify error margins; 2) Test the method on more athletes, varying speeds, and different track conditions to assess robustness; 3) Explore adaptations for non-track settings (e.g., cross-country running) by integrating SLAM or depth sensors; 4) Optimize the pipeline further for edge devices (e.g., smartphones) to enhance accessibility for coaches.

\bibliographystyle{unsrt}  
\bibliography{template}


\end{document}